\documentclass[runningheads]{llncs}
\usepackage[T1]{fontenc}
\usepackage{graphicx}

\usepackage{wrapfig}
\usepackage{multirow}
\usepackage{adjustbox}
\usepackage{pifont}
\usepackage{svg}
\usepackage{subfig}
\usepackage{hyperref}
\newcommand{\cmark}{\ding{51}}%
\newcommand{\xmark}{\ding{55}}%

\newcommand{\orcid}[1]{\href{https://orcid.org/#1}{\includegraphics[width=10pt]{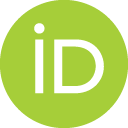}}}

\usepackage{color}

\begin{document}
\title{LiFT: Lightweight, FPGA-tailored 3D object detection based on LiDAR data}
\titlerunning{LiFT: Lightweight, FPGA-tailored 3D object detection}
%
\author{Konrad Lis\inst{1}\orcid{0000-0003-2034-0590} \and
Tomasz Kryjak\inst{1}\orcid{0000-0001-6798-4444} \and
Marek Gorgoń\inst{1}\orcid{0000-0003-1746-1279}}

\authorrunning{K. Lis et al.}

\institute{Embedded Vision Systems Group, \\
   Department of Automatic Control and Robotics, \\
   AGH University of Krakow,  \\
   Al. Mickiewicza 30, 30-059 Krakow, Poland \\ \email{\{kolis,tomasz.kryjak,mago\}@agh.edu.pl}}

\maketitle              
\begin{abstract}

This paper presents LiFT, a~lightweight, fully quantized 3D object detection algorithm for LiDAR data, optimized for real-time inference on FPGA platforms. 
Through an in-depth analysis of FPGA-specific limitations, we identify a~set of FPGA-induced constraints that shape the algorithm's design. 
These include a~computational complexity limit of 30 GMACs (billion multiply-accumulate operations), INT8 quantization for weights and activations, 2D cell-based processing instead of 3D voxels, and minimal use of skip connections.
To meet these constraints while maximizing performance, LiFT combines novel mechanisms with state-of-the-art techniques such as reparameterizable convolutions and fully sparse architecture. 
Key innovations include the Dual-bound Pillar Feature Net, which boosts performance without increasing complexity, and an efficient scheme for INT8 quantization of input features.
With a~computational cost of just 20.73 GMACs, LiFT stands out as one of the few algorithms targeting minimal-complexity 3D object detection. 
Among comparable methods, LiFT ranks first, achieving an mAP of 51.84\% and an NDS of 61.01\% on the challenging NuScenes validation dataset.
%
The code will be available at \url{https://github.com/vision-agh/lift}.

\end{abstract}

\section{Introduction}
\label{sec:introduction}


With the rapid development of autonomous vehicle technology, one of the key challenges becomes ensuring reliable perception of the surroundings. 
In particular, real-time, high-precision 3D objects detection is the foundation for the safety and efficiency of autonomous systems.
Among the various sensors used in autonomous vehicles, LiDAR (Light Detection and Ranging) has gained considerable popularity due to its ability to generate accurate 3D maps of the environment, allowing for precise object detection and classification.
Thanks to its independence from lighting conditions and its high resolution and accuracy, the technology offers advantages over other sensors such as cameras and radars.


Modern 3D object detection algorithms based on LiDAR data typically use Deep Convolutional Neural Networks (DCNNs), which offer high performance but come with significant computational and memory demands.
As a~result, they are usually implemented on high-performance computers with GPU (Graphics Processing Unit) cards for efficient training and inference.
However, the overarching goal of these algorithms is real-time operation in autonomous vehicles or advanced driver assistance systems (ADAS).
For application in an actual, mass-produced vehicle, a~reliable, power-efficient and low-cost computing platform is required -- for example a~modern SoC FPGA.
The real-time reimplementation of SoTA (State of The Art) algorithms on such platforms is a~major challenge and often requires a~significant redesign of the algorithm to take full advantage of the capabilities of the platform under consideration (so-called hardware aware algorithm design).


In this paper, we present LiFT, a~lightweight 3D detector based on LiDAR sensor data, carefully designed to run in real-time on low-power FPGA or ASIC platforms while providing high detection performance on the demanding NuScenes dataset by combining a~number of novel solutions with SoTA mechanisms. 

The main contributions of this paper are:
\begin{itemize}
    \item a~set of constraints on a~3D detector architecture in the context of implementation on an FPGA platform,
    \item an efficient way to quantize initial features,
    \item Dual-Bound Pillar Feature Net -- a~Pillar Feature Net extension to increase detection performance without added complexity,
    \item a~3D detection algorithm adapted for implementation in FPGAs, with the best detection efficiency among comparable methods on the NuScenes dataset.
\end{itemize}

The reminder of this paper is organised as follows.
In Section \ref{sec:related} we discuss issues related to our work: most commonly used datasets, DCNN approaches to object detection in LiDAR data and FPGA/ASIC implementations of such algorithms.
Next, in Section \ref{sec:ha_design} we elaborate on the LiFT design, focusing on hardware induced constraints on the algorithm and novel mechanisms.
The results obtained are summarised in Section \ref{sec:experiments}.
The paper ends with a~short summary with conclusions and discussion of possible future work.

\section{Related work}
\label{sec:related}

\subsubsection{Datasets}
\label{sec:related:3d_obj_det:datasets}

The most commonly used datasets for object detection based on LiDAR data are KITTI\cite{dataset:kitti} (2012), NuScenes\cite{dataset:nuscenes} (2019) and Waymo\cite{dataset:waymo} (2019).
The latter two sets are much larger and more challenging than KITTI.
NuScenes contains 1,000 sequences that add up to 1.4 million images, 390,000 LiDAR scans and 1.4 million tagged objects.
Of the 390k LiDAR scans, only 40k are labelled -- 28310 are used for training, 6019 for validation, and 6008 for testing.
3D object detection on the NuScenes is evaluated using the standard mAP metric (mean Average Precision) and a~metric called NDS (nuScenes detection score).
It includes mAP and several error measures, e.g. orientation error or scale error.


\subsubsection{Methods}
\label{sec:related:3d_obj_det:methods}

There are several approaches to 3D object detection from LiDAR data, the most popular of which are point-based and cell-based methods (the later most common in SoTA solutions).
In the first step, the point cloud is reduced to a~regular grid of 2D or 3D cells (pillars or voxels).
The cells are assigned a~feature vector using the Pillar Feature Encoder (PFE) in the 2D case or the Voxel Feature Encoder (VFE) in the 3D case.
Due to the rationale described in Section \ref{sec:ha_design:constraints}, the review is limited to detectors using 2D cells.


The first and the simplest solution is PointPillars\cite{algorithm:pointpillars}, with the PFE called Pillar Feature Net (PFN), consisting of a~linear layer, Batch Normalization and ReLU, which are applied to each point individually.
Points from each pillar are reduced to a~feature vector using pointwise \textit{Max pooling}.
Another 2D detector is CenterPoint-Pillar\cite{algorithm:centerpoint}, which uses PointPillars as a~backbone and a~novel CenterHead.
In the one-stage version, it predicts a~heatmap with the centres of objects and regression maps specifying their location, shape and orientation.
Both PointPillars and CenterPoint-Pillar are based on regular convolutional neural networks.
However, usually 90\% of the input pillars are empty and are implicitly assigned a~zeroed feature vector, what can be exploited using sparse convolutional neural networks to significantly reduce computations.
In addition to standard sparse layers, which function similarly to regular convolutions by gradually expanding the area of non-zero, so-called active, pixels, there are also submanifold versions of these layers, which maintain the active pixels structure intact.
Other popular 2D detectors, representing the SoTA on datasets such as NuScenes and Waymo, include PillarNet\cite{algorithm:pillarnet} and its derivatives VoxelNeXt-2D\cite{algorithm:voxelnext} and FastPillars\cite{algorithm:fastpillars}.
VoxelNeXt-2D is particularly notable as it incorporates only sparse convolutions, even in its head.


\subsubsection{FPGA/ASIC-based implementations}
\label{sec:related:3d_obj_det:fpga_methods}


To date, there have been few implementations of LiDAR sensor-based object detectors on FPGA platforms, all of which are based on the original or modified version of PointPillars \cite{other:vitis_ai_3_0_model_zoo,lidar_implementation:pointpillars_stanisz,lidar_implementation:pointpillars_vea,lidar_implementation:pointpillars_latotzke,lidar_implementation:pointpillars_brum}.
Most of them operate in real-time, defined as processing 10 point clouds per second, which corresponds to the typical LiDAR rotation frequency.
All of them are evaluated on the KITTI dataset.
However, methods capable of performing well on more challenging datasets, such as the NuScenes are currently missing.


The authors of SPADE\cite{asic_implementation:spade} and SPADE+\cite{asic_implementation:spade_plus} undertook the task of developing a~2D sparse convolution accelerator targeting an ASIC, which was evaluated in simulation.
In addition, they implemented PointPillars, CenterPoint-Pillar and PillarNet in several different sparse versions and tested the accelerator performance.
PointPillars was evaluated on KITTI, while CenterPoint-Pillar and PillarNet were evaluated on NuScenes.


\section{Hardware-aware design of LiFT}
\label{sec:ha_design}

\subsection{Hardware-induced constraints on algorithm design}
\label{sec:ha_design:constraints}


Our goal is to develop a~3D detection algorithm with relatively high detection accuracy, suitable for real-time implementation on an FPGA platform.
As a~reference platform, we will use the AMD/Xilinx Kria K26 SOM platform, a~mid-range SoC FPGA device, which could serve as an embedded processing platform for LiDAR data. 
In the design process, it is essential to consider the constraints imposed by the chosen computational platform, particularly in terms of computational complexity, memory usage, and other algorithmic factors.
It is important to emphasize that the algorithm will not be restricted to this specific platform and could be adapted for use on other FPGA or ASIC devices as well.


The first issue to address, when considering computational complexity, is the precision of the calculations. 
Floating-point operations are more resource-intensive and time-consuming, so, whenever possible and without significant loss of accuracy, it is recommended to perform quantization of the algorithm's computations. 
The most common approach is 8-bit integer quantization (INT8), which, when combined with quantization aware training, provides sufficient detection accuracy with high processing speed and low memory consumption. 
In addition, INT8 quantization is widely supported by most FPGA accelerators and other embedded systems. 
For these reasons, this work opts to use INT8 quantization.


To determine the upper limit on the computational complexity of the 3D detection algorithm, we refer to the DNN accelerator provided by AMD/Xilinx - the DPU (Deep Learning Processor Unit). 
The computational complexity of the neural network will be expressed in terms of MAC (Multiply And Accumulate) operations, representing the number of multiply-accumulate operations required to produce the network's output.
The K26 SOM platform's logic resources can accommodate a~single instance of the B4096 version of the DPU, operating at a~clock frequency of 300 MHz. 
This means that the DPU can perform 2048 MAC operations per clock cycle, resulting in a~processing rate of 614.4 GMAC/s ($1GMAC = 10^{9}MAC$) at a~clock speed of 300 MHz.
Assuming a~typical real-time processing definition, i.e., 10 point clouds per second (pcd/s), the number of operations required for a~single point cloud, in the ideal case, must not exceed 61.44 GMAC. 
In practice, however, determining the exact upper limit for number of operations is not trivial, as all the algorithms include operations that are not included in the assessment of computational complexity.
According to \cite{other:vitis_ai_3_0_model_zoo}, among the algorithms capable of processing 10 frames per second, the maximum observed complexity is 50 GMAC, an average is around 30 GMAC.
Adopting a~limit of 50 GMAC could be overly optimistic, given the potential overheads of LiFT such as dividing point cloud into 2D cells, therefore we assume an upper bound on computational complexity of 30 GMAC. 
Of course, the lower the computational complexity of the algorithm, the better. 
However, reducing complexity is not always justified if it leads to a~significant drop in detection performance.
Ultimately, the algorithm's runtime should be verified on the target platform.
The adopted upper limit of 30 GMAC is merely an approximation, indicating which 3D detection algorithms, operating on LiDAR data, are worth implementing in an embedded system if real-time processing is to be achieved.

When addressing the issue of memory complexity, it is important to first consider the specifics of FPGA systems.
For AMD/Xilinx FPGAs from the Zynq Ultrascale+ series, the system includes both programmable logic with a~small amount of dedicated memory -- referred to as on-chip memory (OCM), typically up to a~few MiB (SRAM), and a~processing system with external DRAM memory of several GiB.
In data processing within programmable logic, it is preferable to use on-chip memory due to its short and predictable access time, which amounts to only a~few clock cycles.
In contrast, operations involving DRAM take a~non-deterministic number of cycles -- tens or more, and they consume significantly more energy compared to operations performed on OCM.


In the context of designing 3D object detectors, the limited amount of on-chip memory becomes a~critical factor influencing the structure of the entire algorithm.
Specifically, two main aspects of the detector design are particularly sensitive to this limitation.
%
%
The first is the amount of skip connections, which require storing entire data tensors.
Due to the small capacity of the OCM, tensor often needs to be stored in the external DRAM.
This results in increased detector latency because of additional data transfers, especially when the number of skip connections is large.
Thus, it is recommended to use architectures with no skip connections or only a~very limited number of them.


The second aspect is the dimensionality of processed cells.
DNN accelerators on FPGA, for each convolutional layer, must perform the \textit{Im2Col} operation, which generates the convolutional context.
It requires buffering the pixels that constitute subsequent contexts, and for efficiency, OCM is preferred for this task.
In the case of a~dense 3D tensor with dimensions XY of 640 × 720 and a~Z-height of 40, assuming a~context size of 3x3x3, a~buffer of $40 \times 640 \times 2 + 640 \times 2 + 3 = 52483$  cells is required.
In comparison, for a~2D tensor with the same XY dimensions and a~context size of 3x3, only $640 \times 2 + 3 = 1283$ cells are needed -- over 40x fewer than in the 3D case.
Given the limited on-chip memory, it is therefore advisable to use detectors operating on 2D cells rather than 3D cells.


In summary, the computational platform and the requirement for real-time processing impose several constraints on the detection algorithm.
The limited amount of on-chip memory necessitates operating on a~two-dimensional grid of cells and restricting the number of skip connections to a~maximum of a~few.
Meanwhile, considerations related to the computational power and logical resources of the FPGA mandate the use of INT8 quantization and limit the computational complexity to 30 GMAC.

\subsection{Constraints analysis and design assumptions}
\label{sec:ha_design:assumptions}


When looking for an architecture that satisfies the constraints described in the Section \ref{sec:ha_design:constraints}, the detection efficiency should be maximised at the same time.
All of the aforementioned limitations, in the general case, result in a~reduction of the precision and amount of information available in the detector or potentially a~less efficient learning process.
In this subsection we analyse adopted constraints and establish additional assumptions to achieve a~high performance detector.


Skip connections are widely used in neural network architectures to address the vanishing gradient problem during training and to combine features across different scales.
Recently, the RepVGG approach \cite{algorithm:repvgg} has been introduced, allowing the use of skip connections spanning across single layers during training while removing them during inference through a~simple weight transformation.
In designing our computational architecture, we will leverage the RepVGG mechanism to eliminate most skip connections.
However, we will retain residual connections related to multi-scale feature fusion, as their removal would result in a~significant drop in detection accuracy, as demonstrated in Section \ref{sec:experiments:ablation}.


When processing data as 2D cells instead of 3D cells, constructing an effective feature vector for each cell becomes more challenging, as all dimensions of input data must be preserved due to detection in three dimensions.
For 3D cells a~common approach is to average point features in a~given cell, however it would result in a~significant height information loss in the 2D cells case.
For this reason, more advanced solutions are used.
The most common encoder is PFN from \cite{algorithm:pointpillars}, which is used in many other detectors such as \cite{algorithm:pillarnet} and \cite{algorithm:voxelnext}, or its extensions, such as MAPE from \cite{algorithm:fastpillars}.
In our architecture, we will use a~custom extension of PFN -- DBPFN -- described in more detail in Section \ref{sec:ha_design:dual_bound_pfn}.
By employing \textit{Min pooling} alongside \textit{Max pooling}, our solution minimally complicates the PFN computational architecture while significantly increasing its effectiveness.


Another important aspect to analyse is computation quantization.
We intend to use INT8 quantization, so it is crucial that the input features are well represented in 8 bits to avoid losing too much information at the beginning.
Thanks to the method described in Section \ref{sec:ha_design:input_quantisation}, we leverage both the convenience of using 8-bit quantisation everywhere in LiFT and the effective localization resolution below 2 mm compared to 40 cm in a~default quantisation case.


Considering the computational complexity and referring to contemporary 3D detectors, we plan to use sparse convolutions.
These reduce computational complexity by an average of 50\% to 80\%, depending on the specific architecture.
However, achieving this reduction entails an additional cost due to the more complex context generation process.


In conclusion, having analysed the constraints from Section \ref{sec:ha_design:constraints}, we propose using short skip connections as introduced in RepVGG, encoding features for pillars with DBPFN, implementing efficient input feature quantization, and utilizing sparse convolutions.

\subsection{Dual-Bound Pillar Feature Net}
\label{sec:ha_design:dual_bound_pfn}


The structure of the original PFN is simple and relatively efficient.
It is commonly used, e.g. in PillarNet\cite{algorithm:pillarnet} and VoxelNeXt\cite{algorithm:voxelnext}.
There is potential to increase its efficiency through various extensions, but this comes with the added complexity of the architecture, for instance, by introducing additional operation like self-attention as seen in MAPE\cite{algorithm:fastpillars}. 
In our work, we propose an extended version of PFN, designed in such a~way as to preserve its simplicity. 
This allows us to avoid complicating the FPGA implementation by introducing new operators.

\begin{figure}[t]
\includegraphics[width=\textwidth]{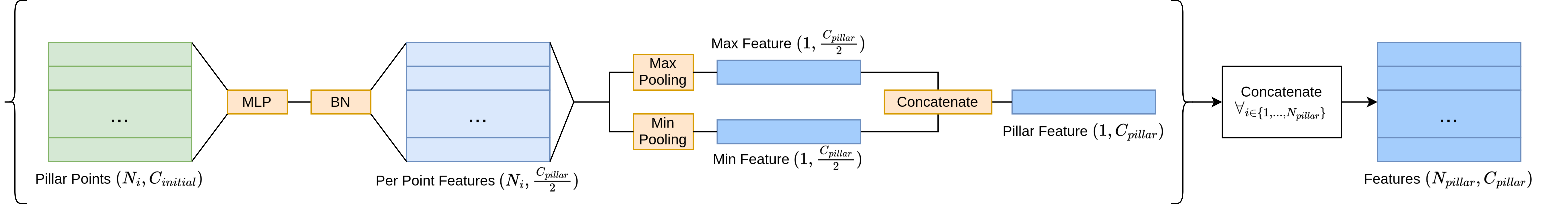}
\caption{An outlook on Dual-Bound Pillar Feature Net (DBPFN) structure}
\label{fig:ha_design:dual_bound_pfn}
\end{figure}


The PFN, despite its relative efficiency and popularity, has several downsides.
One is the significant reduction in information about the distribution of points in the pillar.
Features from all points are reduced to a~single vector by the \textit{Max pooling} operation, thus potentially losing most of the information about the distribution of features within the pillar. 
Another issue is the blockage of gradient flow to the MLP due to the \textit{ReLU} activation function and \textit{Max Pooling}.
Through the \textit{Max Pooling} operation, each output feature from the MLP is only learned based on one point within the given pillar -- specifically, the one that happened to have the largest value of the feature.
Additionally, if the data and weight distributions cause the activation entering \textit{ReLU} to be negative in most cases, the neuron will fall into the region of \textit{ReLU} with a~zero slope. This results in a~zero gradient for that feature in the MLP, leading to what is known as a~``dead neuron'' which will stop getting updates during training.


These issues can be partially addressed by a~small modification to PFN that only slightly complicates the architecture -- removing the \textit{ReLU} activation function, using \textit{Min Pooling} alongside \textit{Max Pooling}, and concatenating the features obtained from both operations. 
We call this modified PFN the Dual-Bound Pillar Feature Net (DBPFN), and its structure is shown in Figure \ref{fig:ha_design:dual_bound_pfn}.
By incorporating \textit{Min Pooling} in addition to \textit{Max Pooling}, the individual MLP features are learned from two points within the pillar instead of just one, preserving more information about the feature distribution inside the pillar. 
This is not a~major quantitative change, but selecting the minimum point alongside the maximum potentially diversifies the set of points on which each MLP feature is learned.
With the addition of \textit{Min Pooling}, removing \textit{ReLU} becomes essential, as \textit{ReLU} would zero out negative values, causing \textit{Min Pooling} to return zeros more often instead of more descriptive feature values.
%
%
To maintain the same number of output features in the DBPFN as in the original implementation, the number of features in the MLP must be reduced by a~half. 
However, as it will be shown in Section \ref{sec:experiments:ablation}, our implementation is still more efficient than the original PFN.

\subsection{Effective input features quantisation}
\label{sec:ha_design:input_quantisation}



We opt to apply INT8 quantization, so in order to minimize information loss, the input features must be well represented with 8 bits. 
Detectors working on the NuScenes dataset typically use a~range of -54m to 54m in both the X~and Y~dimensions.
Assuming we want to utilize the full range, we would achieve a~resolution of about 40 cm with 8-bit quantization. 
This is far too coarse compared to the size of the pillars, which is 15 cm x~15 cm in our case. 
For this reason, we divided each of the XYZ dimensions into two features: a~coarse and a~detailed location.
For the X~dimension, the definitions of the features $X_{coarse}$ and $X_{detail}$ are as follows:
\[X_{coarse} = \left\lfloor \frac{X}{resolution_{X}} \right\rfloor \times resolution_{X} \quad \quad
X_{detail} = X~- X_{coarse}\]
where $resolution_{X} = 2^{-8} * \left(X_{max} - X_{min}\right)$, $X$ denotes the location of a~given point along the X-axis, and $X_{min}$ and $X_{max}$ define the boundaries of the point cloud along the same axis. 
In a~similar manner, the features $Y_{coarse}$, $Y_{detail}$, $Z_{coarse}$, and $Z_{detail}$ are defined.

\subsection{LiFT design}
\label{sec:ha_design:design}

\begin{figure}[t]
\centering
\includegraphics[width=\textwidth]{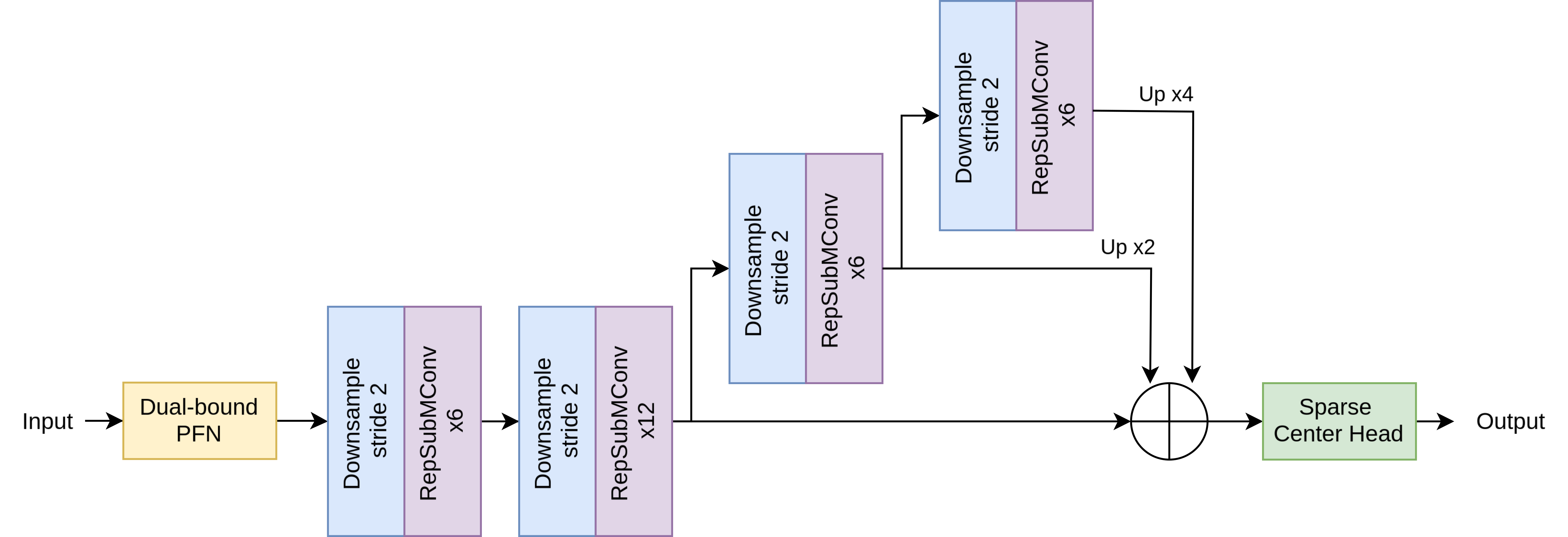}
\caption{An outlook on LiFT structure}
\label{fig:ha_design:lift_scheme}
\end{figure}

The schematic of our proposed architecture -- LiFT -- is shown in Figure \ref{fig:ha_design:lift_scheme}.
The detector operates on 2D cells and consists entirely of sparse convolutions, including the head.
All weights and activations are quantized with type INT8.
We have also used the effective quantization of input features described in Section \ref{sec:ha_design:input_quantisation}.
As the PFE, we use DBPFN, described in Section \ref{sec:ha_design:dual_bound_pfn}.


\begin{figure}[b]
\centering
\includegraphics[width=0.5\textwidth]{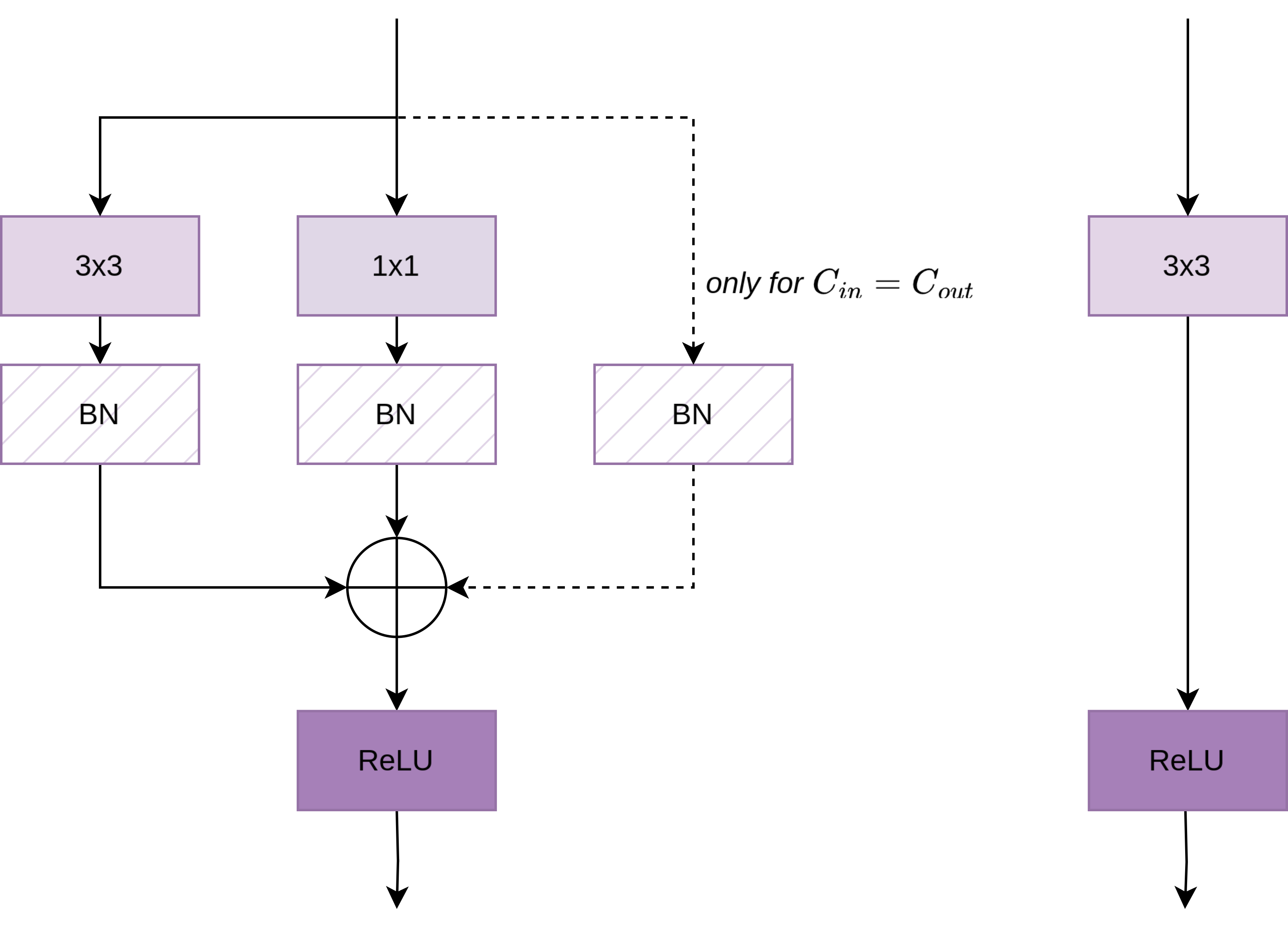}
\caption{Reparametrisable convolution structure during training (on the left) and during inference (on the right).}
\label{fig:ha_design:rep_conv}
\end{figure}

To remove short skip connections during inference, we use a~reparameterisation scheme from RepVGG.
Thus, we define an ordinary reparametrisable sparse convolution layer -- \textit{RepSparseConv} and its submanifold version \textit{RepSubMConv}.
Their structure during training and inference is shown in Fig. \ref{fig:ha_design:rep_conv}.
In the backbone, we use 4 stages -- each stage consisting of a~downsampling layer based on RepSparseConv with a~kernel size of 3x3 and stride equal 2, and a~number of RepSubMConv layers with a~kernel size of 3x3 and stride equal 1.
The number of RepSubMConv layers in each stage is 6, 12, 6, and 6, respectively.


When it comes to fusing features from multiple scales, we use an approach inspired by the one applied in VoxelNeXt 2D\cite{algorithm:voxelnext}.
This involves adding the output tensors from the 3rd and 4th stages to the tensor from the 2nd stage, previously upsampling them to the resolution of the 2nd stage.
In VoxelNeXt 2D, tensor upsampling from the 3rd and 4th stages consisted of simply multiplying the coordinates of the sparse tensor by 2 and 4, respectively.
However, this procedure created only one pixel while the resolution changed 2x or 4x.
Moreover, in general case, it did not align with the corresponding active pixels in the second stage.
Therefore, we introduced a~different strategy for changing the resolution.
When adding tensors, we keep the active pixels from the second stage only and add the corresponding pixels from the third and fourth stages to them.
Our solution is more efficient than the VoxelNeXt's one, as it will be shown in Section \ref{sec:experiments:ablation}.
The head design, on the other hand, is borrowed from the VoxelNeXt -- it is a~Sparse Center Head based entirely on sparse convolutions.

\section{Experiments}
\label{sec:experiments}


The network was trained on the NuScenes dataset.
We use a~pillar size of 15cm x~15cm and a~point cloud range of [-54m, 54m] along the X~and Y~axes and [-5m, 3m] along the Z~axis.
The number of features in the PFE output is 64, while in each respective stage is 64, 64, 128, 128.
In the output of the second stage, we apply an additional layer to align the number of features with the 3rd an the 4th stage.
When it comes to data augmentation and training hyperparameters, we follow the VoxelNeXt settings.
We ran the training on a~computer with 7 Nvidia RTX A6000 cards. 
The total batch size was equal to 42.



\subsection{Overall results}
\label{sec:experiments:overall}

LiFT evaluation results are presented in Table \ref{tab:results:overall}.
It was compared with other detectors evaluated on the NuScenes, which meet the constraints defined in the Section \ref{sec:ha_design:constraints}.
The only detectors of this kind, except for LiFT, are different sparse versions of the CenterPoint, presented in \cite{asic_implementation:spade} and \cite{asic_implementation:spade_plus}.
To the best of our knowledge, no other work has presented a~detector that meets the aforementioned constraints.
Nor are we aware of any other detector with a~complexity of less than 30 GMAC, even after relaxing the other limitations.
LiFT achieved 51.84\% mAP and 61.01\% NDS on the NuScenes-val set with a~computational complexity of 20.73 GMAC.
In terms of mAP and NDS metrics, it ranks first, with a~large margin of 9.27 GMAC from the computational complexity threshold.


\begin{table}[t]
\centering
\caption{Performance comparison of 3D object detection algorithms based on 2D cells with number of operations equal at most 30 GMAC.}
\label{tab:results:overall}
\begin{tabular}{c|l|l|l|l}
\hline
Source                  & \multicolumn{1}{c|}{Detector} & \multicolumn{1}{c|}{mAP {[}\%{]}} & \multicolumn{1}{c|}{NDS {[}\%{]}} & \multicolumn{1}{c}{GMAC} \\ \hline
ours                    & \textbf{LiFT}                          & \textbf{51.84}                    & \textbf{61.01}                    & 20.73                    \\ \hline
\multirow{2}{*}{SPADE\cite{asic_implementation:spade}}  & SCP2                          & 50.12                             & 60.42                             & 24.77                    \\
                        & SCP3                          & 47.78                             & 58.97                             & \textbf{13.60}           \\ \hline
\multirow{4}{*}{SPADE+\cite{asic_implementation:spade_plus}} & SparseCenterPoint - SubM-Conv & 47.89                             & 58.94                             & 18.13                    \\
                        & SparseCenterPoint - PS-Conv   & 50.12                             & 60.42                             & 27.09                    \\
                        & SparseCenterPoint - FS-Conv   & 50.30                             & 60.41                             & 23.34                    \\
                        & SparseCenterPoint - SD-Conv   & 50.33                             & 60.84                             & 19.37                    \\ \hline
\end{tabular}
\end{table}

Relative to the second best solution, \textit{SparseCenterPoint - SD-Conv} from \cite{asic_implementation:spade_plus}, LiFT is better by 1.51\% in terms of the mAP metric and 0.17\% in terms of NDS, respectively.
At the same time, LiFT has a~slightly higher computational complexity -- by 1.36 GMAC.
As we mentioned in Section \ref{sec:ha_design:constraints}, even with real-time processing achieved, reducing computational complexity is still desirable.
However it can not result in a~significant drop of mAP and NDS metric.
A~potential 1.36 GMAC reduction in computational complexity, obtained by choosing the second best solution, could be justified in terms of the NDS metric - the decrease is only 0.17\%.
In contrast, the mAP metric shows a~much more considerable decline of 1.51\%.
For this reason, we believe that according to the constraints and assumptions made, LiFT is the best choice for implementation on an FPGA platform.

To conclude the discussion and provide a comprehensive comparison with SoTA methods, we now focus exclusively on detection performance, disregarding all other factors.
Although LiFT falls significantly short of SoTA detection accuracy under unconstrained conditions, it still outperforms a considerable 12\% of all submissions (in January 2025) on the NuScenes\cite{dataset:nuscenes} leaderboard, including PointPillars\cite{algorithm:pointpillars} with mAP of 30.5\% and NDS equal to 45.3\%.


\subsection{Ablation studies}
\label{sec:experiments:ablation}

In the ablation study, we analysed the impact of removing multi-scale processing as well as four distinct components of LiFT that are crucial to its high detection performance.
The result of experiments, proving the effectiveness of the four incorporated solutions, are presented in Table \ref{tab:results:ablation}.
We compared the DBPFN against the implementation with the original PFN -- of the components analysed, it has the greatest impact on detection efficiency.
The effective quantization of the initial features was challenged against a~naive solution in which none of the features are split and all are quantized with INT8 type.
This element has a~similar impact on detection performance as the introduction of reparametrisable convolutions.
We compared their efficiency against architectures with the same number of layers, but based on a~residual connection structure called BasicBlock, commonly used in 3D detectors, such as \cite{algorithm:voxelnext}\cite{algorithm:pillarnet}.
It turns out that the introduction of reparametrizable convolutions instead of classical skip connection structures does not only allow to obtain a~simple structure during inference, but also positively affects the detection efficiency.
The last element examined is the effective scale fusion described in Section \ref{sec:ha_design:design}, which was compared against an implementation from VoxelNeXt 2D.
Our solution has slightly better detection performance while as simple as the one from VoxelNeXt.
In addition, it allowed us to make a~small but noticeable decrease in computational complexity by reducing the number of pixels that enter the Sparse CenterHead.

\begin{table}[t]
\centering
\caption{Effects of different components of LiFT on mAP and NDS}
\label{tab:results:ablation}
\begin{tabular}{c|c|c|c|l|l}
\hline
\begin{tabular}[c]{@{}c@{}}Efficient\\ scale fusion\end{tabular} & \begin{tabular}[c]{@{}c@{}}Reparametrisable\\ convolutions\end{tabular} & \begin{tabular}[c]{@{}c@{}}Efficient\\ quantisation\end{tabular} & \begin{tabular}[c]{@{}c@{}}Dual-bound\\ PFN\end{tabular} & \multicolumn{1}{c|}{mAP {[}\%{]}} & \multicolumn{1}{c}{NDS {[}\%{]}} \\ \hline
\cmark                                            & \cmark                                                   & \cmark                                            & \cmark                                    & \textbf{51.84}                            & \textbf{61.01}                           \\ \hline
\xmark                                            & \cmark                                                   & \cmark                                            & \cmark                                    & 51.49 (-0.35)                             & 60.73 (-0.28)                            \\ \hline
\cmark                                            & \xmark                                                   & \cmark                                            & \cmark                                    & 51.05 (-0.79)                             & 60.63 (-0.38)                            \\ \hline
\cmark                                            & \cmark                                                   & \xmark                                            & \cmark                                    & 51.13 (-0.71)                             & 60.47 (-0.54)                            \\ \hline
\cmark                                            & \cmark                                                   & \cmark                                            & \xmark                                    & 50.29 (-1.55)                             & 59.87 (-1.14)                            \\ \hline
\end{tabular}
\end{table}

We conducted an additional experiment in which we removed multiscale processing from LiFT.
Detection efficiency, as measured in terms of both mAP and NDS, dropped significantly -- by 9.43\% and 6.69\%, respectively.
Potentially, removing the last two skip connections would allow a~slight speed-up in network performance by removing a~few transfers between the FPGA and the external memory and reducing external memory consumption.
However, the gain is so small that it is not justified with such a~large decrease in efficiency, so we decided to keep the multiscale processing.



\section{Conclusion and Discussion}
\label{sec:conclusion}

In this paper, we present \textbf{LiFT}, a~3D object detection algorithm tailored for real-time implementation on FPGA. 
LiFT has been carefully designed, starting with determining the hardware induced constraints.
In the next step we introduced Dual-Bound Pillar Feature Net and an efficient scheme to quantize INT8 input features so as to provide high detection performance while fulfilling the limitations.
By combining a~holistic analysis of the problem with our novel mechanisms and SoTA solutions, such as reparametrizable convolutions and fully sparse architecture, we obtained an algorithm with the best detection performance among methods of comparable complexity.
We achieved 51.84\% mAP and 61.01\% NDS on the NuScenes-val set with a~computational complexity of 20.73 GMAC.
We believe we have developed a~solid baseline for implementation on FPGAs and we hope our work will encourage the research community to pursue the topic of designing 3D detectors for embedded devices more often.


In future work, we plan to implement LiFT on FPGA.
In addition, we plan to further improve the LiFT architecture, e.g. by using Spatially-Dilated Sparse Convolutions from SPADE+, which, according to the authors, significantly increase detection efficiency due to an increased reception field with little computational complexity overhead.


\begin{credits}

\subsubsection{\discintname}
The authors have no competing interests to declare that are relevant to the content of this article.

\subsubsection{\ackname} The work presented in this paper was supported by the AGH University of Krakow, project no. 10.16.120.79990 and the program ``Excellence initiative -- research university'' for the AGH University of Krakow.


\end{credits}
%
%
%
\bibliographystyle{splncs04}
\bibliography{algorithms, asic_implementations, datasets, libraries, lidar_implementations, others}

\begin{thebibliography}{10}
\providecommand{\url}[1]{\texttt{#1}}
\providecommand{\urlprefix}{URL }
\providecommand{\doi}[1]{https://doi.org/#1}

\bibitem{other:vitis_ai_3_0_model_zoo}
AMD/Xilinx: Vitis ai model zoo.
  \url{https://xilinx.github.io/Vitis-AI/3.0/html/docs/workflow-model-zoo.html}
  (Last access 22th November 2024)

\bibitem{lidar_implementation:pointpillars_brum}
Brum, H., V{\'e}stias, M., Neto, H.: Lidar 3d object detection in fpga
  with low bitwidth quantization. In: Applied Reconfigurable Computing.
  Architectures, Tools, and Applications. pp. 90--105. Springer Nature
  Switzerland, Cham (2024)

\bibitem{dataset:nuscenes}
Caesar, H., Bankiti, V., Lang, A.H., Vora, S., et~al.: nuscenes: A multimodal
  dataset for autonomous driving. arXiv preprint arXiv:1903.11027  (2019)

\bibitem{algorithm:voxelnext}
Chen, Y., Liu, J., Zhang, X., Qi, X., et~al.: Voxelnext: Fully sparse voxelnet
  for 3d object detection and tracking. In: Proceedings of the IEEE/CVF
  Conference on Computer Vision and Pattern Recognition (CVPR). pp.
  21674--21683 (June 2023)

\bibitem{algorithm:repvgg}
Ding, X., Zhang, X., Ma, N., Han, J., et~al.: Repvgg: Making vgg-style convnets
  great again. In: Proceedings of the IEEE/CVF Conference on Computer Vision
  and Pattern Recognition (CVPR). pp. 13733--13742 (June 2021)

\bibitem{dataset:kitti}
Geiger, A., Lenz, P., Stiller, C., Urtasun, R.: Vision meets robotics: The
  kitti dataset. International Journal of Robotics Research (IJRR)  (2013)

\bibitem{algorithm:pointpillars}
{Lang}, A.H., {Vora}, S., {Caesar}, H., et~al., L.Z.: Pointpillars: Fast
  encoders for object detection from point clouds. In: 2019 IEEE/CVF Conference
  on Computer Vision and Pattern Recognition (CVPR). pp. 12689--12697 (June
  2019)

\bibitem{lidar_implementation:pointpillars_latotzke}
Latotzke, C., Kloeker, A., Schoening, S., Kemper, F., et~al.: Fpga-based
  acceleration of lidar point cloud processing and detection on the edge. In:
  2023 IEEE Intelligent Vehicles Symposium (IV). pp.~1--8 (2023)

\bibitem{asic_implementation:spade}
Lee, M., Park, S., Kim, H., Yoon, M., et~al.: Spade: Sparse pillar-based 3d
  object detection accelerator for autonomous driving. In: 2024 IEEE
  International Symposium on High-Performance Computer Architecture (HPCA). pp.
  454--467 (2024)

\bibitem{lidar_implementation:pointpillars_vea}
Li, X., Ren, A., Tan, Y., Li, X., et~al.: Vea: An fpga-based voxel encoding
  accelerator for 3d object detection with lidar. In: 2022 IEEE 40th
  International Conference on Computer Design (ICCD). pp. 509--516 (2022)

\bibitem{asic_implementation:spade_plus}
Park, S., Lee, M., Choi, J., Choi, J.: Selectively dilated convolution for
  accuracy-preserving sparse pillar-based embedded 3d object detection. In:
  Proceedings of the IEEE/CVF Conference on Computer Vision and Pattern
  Recognition (CVPR) Workshops. pp. 8104--8113 (June 2024)

\bibitem{algorithm:pillarnet}
Shi, G., Li, R., Ma, C.: Pillarnet: Real-time and high-performance
  pillar-based 3d object detection. In: Computer Vision -- ECCV 2022. pp.
  35--52. Springer Nature Switzerland, Cham (2022)

\bibitem{lidar_implementation:pointpillars_stanisz}
{Stanisz}, J., {Lis}, K., {Gorgon}, M.: Implementation of the pointpillars
  network for 3d object detection in reprogrammable heterogeneous devices using
  finn. Journal of Signal Processing Systems  (2021)

\bibitem{dataset:waymo}
Sun, P., Kretzschmar, H., Dotiwalla, X., Chouard, A., et~al.: Scalability in
  perception for autonomous driving: Waymo open dataset (2019)

\bibitem{algorithm:centerpoint}
Yin, T., Zhou, X., Krahenbuhl, P.: Center-based 3d object detection and
  tracking. In: Proceedings of the IEEE/CVF Conference on Computer Vision and
  Pattern Recognition (CVPR). pp. 11784--11793 (June 2021)

\bibitem{algorithm:fastpillars}
Zhou, S., Tian, Z., Chu, X., Zhang, X., et~al.: Fastpillars: A
  deployment-friendly pillar-based 3d detector (2023),
  \url{https://arxiv.org/abs/2302.02367}

\end{thebibliography}

\end{document}